\documentclass{article} % For LaTeX2e
\usepackage{iclr2025_conference,times}

\usepackage{amsmath,amsfonts,bm}

\def\eqref#1{equation~\ref{#1}}
\def\1{\bm{1}}

\DeclareMathAlphabet{\mathsfit}{\encodingdefault}{\sfdefault}{m}{sl}
\SetMathAlphabet{\mathsfit}{bold}{\encodingdefault}{\sfdefault}{bx}{n}

\usepackage{hyperref}
\usepackage{graphicx}
\usepackage{wrapfig}
\usepackage{url}
\usepackage{booktabs}
\usepackage{multirow}
\usepackage{xspace}
\usepackage{xcolor}
\usepackage{colortbl}

\definecolor{lightgray}{gray}{0.9} % Define the light gray color

\title{LLM-Rank: A Graph Theoretical Approach to Pruning Large Language Models}

\author{David B. Hoffmann, \& Kailash Budhathoki \& Matthaeus Kleindessner \\
Amazon Web Services Research, Germany\\
\texttt{\{adavidho,kaibud,matkle\}@amazon.com} \\
}

\iclrfinalcopy % Uncomment for camera-ready version, but NOT for submission.
\begin{document}
\newcommand{\llmrank}{\textsc{LLM-Rank}\xspace}
\newcommand{\mlprank}{\textsc{MLP-Rank}\xspace}

\maketitle

\begin{abstract}
The evolving capabilities of large language models are accompanied by growing sizes and deployment costs, necessitating effective inference optimisation techniques. We propose a novel pruning method utilising centrality measures from graph theory, reducing both the computational requirements and the memory footprint of these models. Specifically, we devise a method for creating a weighted directed acyclical graph representation of multilayer perceptrons to which we apply a modified version of the weighted PageRank centrality measure to compute node importance scores. In combination with uniform pruning this leads to structured sparsity. We call this pruning method \mlprank. Furthermore we introduce an extension to decoder-only transformer models and call it \llmrank. For both variants we demonstrate a strong performance. With \mlprank on average leading to 6.09 \% higher accuracy retention than three popular baselines and 13.42 \% with \llmrank compared to two popular baselines. 
Code is available at: \url{https://github.com/amazon-science/llm-rank-pruning}. % TODO comment in for camera ready
\end{abstract}

\section{Introduction}
\label{introduction}

%  Context
Over recent years, the increasing popularity of large language models (LLMs), as highlighted by \cite{brown_language_2020}, has driven a trend toward the development of ever-larger and more powerful architectures. This trend has been marked by the regular release of models with growing capabilities, extending from the modelling of language \citep{anthropic_claude_2024, brown_language_2020} to applications in time series forecasting \citep{ansari_chronos_2024}. These transformer-based models typically utilise billions to trillions of parameters \citep{chitty-venkata_survey_2023}. 
\par
% Problem
This results in the problem that the training of such massive models necessitates immense computational power, and their deployment for hosting and inference imposes substantial demands on both compute resources and memory. Moreover, the trend toward larger models is anticipated to persist. 
\par
% Insights
While expanding computing and memory capacity is one approach to support the training and deployment of these models, the field of inference optimisation presents a compelling alternative. By enhancing the efficiency of trained models, inference optimisation reduces the need for additional hardware. This has led to a growing demand for techniques such as pruning, which aims to optimise inference performance by identifying and removing dispensable parameters. We find that the intuitive representation of deep neural networks as graphs makes the use of centrality measures from Graph theory for computing pruning scores an intriguing area of research.  
\par
% Prior work
Although, new state-of-the-art pruning methods such as \cite{ma_llm-pruner_2023, sun_simple_2023, slice2024} are being published regularly, we are not aware of any work directly applying a graph theoretical view to pruning LLMs. The only directly related work we are aware of is \cite{li_pruning_2020}, who apply the Katz centrality measure as a scoring function for node importance in artificial neural networks with the small world property \cite{milgram_small_1967}. Even though not explicitly related to graph theory, the following group of eigenvalue-based pruning methods \cite{buffoni_spectral_2022, cugu_deeper_2022, gkalelis_structured_2020} is relevant due to their similarity to the concept eigenvalue centrality from graph theory. We can generally observe that many zero-order pruning methods although without explicit connection to graph theory, could be rephrased as a network centrality measure when view through a different lens. 
% Soltuion
\par
Making the use of centrality measure concrete, we propose a new method, named \mlprank for inference optimisation of multilayer perceptrons and then extend this method to decoder-only transformer models and call it \llmrank. In particular, this paper proposes the use of the weighted PageRank (WPR) centrality measure from the field of graph theory as a scoring function for structured pruning. This is achieved by generating a weighted directed acyclical graph representation from a neural network model where each node in the graph corresponds to a neuron in the model. The centrality measure is directly applied to this graph representation, computing an importance score for each node corresponding to rows in the models weight matrices. We then use structured uniform pruning to remove entire rows of the model weight matrices based on the computed scores. The resulting structured sparsity does not require specialised hardware to achieve real world inference speedups. 
The use of WPR  \cite{zhang_pagerank_2021} as the underlying centrality measure is due to its design intended for capturing information flow and linked structure in large real-world networks. 
\par
The main contribution of this paper is twofold. Firstly, we propose a way of generating a directed weighted graph representation from both simple multilayer perceptron (MLP) models and more complicated decoder-only transformer models. Secondly, we introduce a variation of WPR centrality which is an adaptation of \cite{zhang_pagerank_2021} to the context of neural network pruning. This leads to the introduction of two novel pruning methods for inference optimisation, namely \mlprank and \llmrank, which we show to have competitive performance to the state-of-the-art structured pruning method SliceGPT \cite{slice2024}, outperforming it with an on average 8.60 \% higher accuracy for the same sparsity ratio.

\section{Preliminaries}
\label{preliminaries}

\subsection{Weighted PageRank}
\label{sec:wpr}

The original PageRank (PR) algorithm was designed to rank web pages as part of the Google search engine \citep{brin_anatomy_1998}. More broadly, PR is a measure determining node centrality in a directed graph $G_d(V, E, \psi)$, where $V$ is the set of vertices, $E$ is the set of edges and $\psi: E \rightarrow \{(v,w)\in V\times V\}$ is a mapping function determining which two vertices an edge connects. The adjacency matrix $A$ is a common way to represent such a graph. $A_{ij} = 1$ if there is an edge from node $V_j$ to $V_i$ and otherwise $A_{ij} = 0$. Each node has an out-degree defined as $d_j^{out} = \sum_l A_{lj}$.  Even though they are not directly related, PR can be conceptualised as an extension of Katz centrality \cite{katz_new_1953}, which itself can be considered a generalisation of the eigenvector centrality measure \cite{bonacich_factoring_1972}. The PR score $\phi_i$ of vertex $V_i$ is recursively defined as:

\begin{equation}
    \phi_i = \gamma \sum_j A_{ij} \frac{\phi_j}{d_j^{out}} + \frac{1-\gamma}{|V|}
    \label{equ:background_weighted_page_rank_pr_definition}
\end{equation}

A variant named WPR was introduced by \cite{zhang_pagerank_2021}. It is created for weighted, directed networks with non-uniform node-specific information $\beta_i$ that can be dependent or independent of the network structure. They propose the WPR measure recursively as defined in \autoref{equ:background_weighted_page_rank_wpr_definition}.

\begin{equation}
    \phi_i = \gamma \sum_j \Big(\theta\frac{W_{ij}}{s_j^{out}} + (1-\theta) \frac{A_{ij}}{d_j^{out}}\Big)\phi_j + \frac{(1-\gamma)\beta_i}{\sum_l \beta_l}
    \label{equ:background_weighted_page_rank_wpr_definition}
\end{equation}

Here, $W$ is the network weight matrix with $W_{ij}$ being the weight associated with the edge from node $V_j$ to $V_i$. The out-strength $s_j^{out}$ is defined as the sum over the outgoing weights of node $j$, namely $s_j^{out} = \sum_l W_{lj}$. Both $\gamma$ and $\theta$ are tuning parameters between zero and one, that control the trade-off between the graph structure and auxiliary node information $\beta_i$ as well as the trade-off between adjacency and weight information, respectively. When $\theta = 0$ and $\beta_i = 1$ WPR is equivalent to PR as introduced by \cite{brin_anatomy_1998} and defined in \autoref{equ:background_weighted_page_rank_pr_definition}.

\subsection{Multilayer Perceptorns}

The MLP is a dense feedforward neural network architecture with multiple layers. The network is considered to be feedforward as information gets sequentially fed through its $K$ densely connected layers. Each layer has several neurons which consist of a linear combination of the layer inputs and a non-linear activation function, as defined in \autoref{equ:mlp_neuron}, where $\bm x \in \mathbb{R}^n$ is the neuron input, $\bm w\in \mathbb{R}^n$ are the weights of the linear transformation, $\bm{wx}$ is the scalar product of $\bm{w}$ and $\bm{x}$, $b \in \mathbb{R}$ is the bias and $\sigma$ is a non-linear activation function such as a sigmoid function or ReLU. 

\begin{equation}
    f(\bm x) = \sigma(\bm{wx} + b)
    \label{equ:mlp_neuron}
\end{equation}

Here $f$ defines the function of a single neuron. Respectively, the entire $k$-th layer of the MLP can be computed by stacking the individual weight vectors to a matrix $\bm W^{(k)} \in \mathbb{R}^{m^{(k)},n^{(k)}}$ and the bias scalars to a vector $\bm b^{(k)} \in \mathbb{R}^{m^{(k)}}$. Here $\bm X^{(k)}$ is input to the $k$-th layer. The layer transformation is then defined as $\bm X^{(k+1)} = \sigma(\bm W^{(k)} \bm X^{(k)} + \bm b^{(k)})$.

\subsection{Decoder-Only Transformers}

Introduced by \cite{vaswani2023attentionneed} as a sequence transduction model, the transformer architecture consists of an encoder and a decoder both with a similar structure. With the popularisation of large language models, a plethora of transformer-based architectures have emerged, such as decoder-only models, which were first introduced by \cite{radford2018improving} and are used for generation tasks. 
\par
Most variants are largely based on the original architecture in which both the encoder and decoder are composed of $n$ blocks. Typically the embedded and positionally encoded input gets fed into the first of these blocks and is in the following incrementally transformed and passed to the next block. The final output is then fed into the final linear layer with softmax activation to predict the next token in the sequence. 
Each of the decoder blocks consists of a masked multi-head attention (MHA) component with three weight matrices for key $K$, value $V$, and query $Q$ projections as defined in \autoref{equ:attention} as well as a position-wise feed-forward network (FFN) with two layers. Both come with skip connections and are followed by a layer norm to facilitate more stable training. 

\begin{equation}
    Attention(Q,K,V) = softmax(\frac{QK^T}{\sqrt{d_k}})V
    \label{equ:attention}
\end{equation}

\section{MLP-Rank Method}

In the following we first propose a way of representing multilayer perceptrons as a graph. We then introduce a modification of the WPR centrality measure as a way of computing pruning importance scores. Together these constitute the \mlprank method. 

\subsection{Graph Representation}
\label{sec:graph_representation}

We propose a simple representation of the MLP as a weighted, directed and acyclical graph. This representation is directly derived from the model's weight matrices, where $W^{(k)}$ corresponds to the  $k$th layer of the neural network and has the dimension $m^{(k)}\times n^{(k)}$. 
In the graph representation, layer $k$ has $m^{(k)}$ vertices  which represent the neurons of the projection $W^{(k)}$. As such, the weight matrix of the graph $W^G$ can be arranged as a block matrix where a partition either consists of an MLP weight matrix or of zeros. 
The trace of blocks directly below the centre diagonal contains the weight matrices of the MLP, which are defined in \autoref{equ:graph_representation} for positive $k$ and visualised in \ref{apx:graph_representation}. 

\begin{equation}
    W^G_{(n^{(0)}+\sum^{k-1}_i m^{(i)}):(n^{(0)}+\sum^{k}_i m^{(i)}), (\sum^{k-1}_i n^{(i)}):(\sum^{k}_i n^{(i)})} = W^{(k)}
    \label{equ:graph_representation}
\end{equation}

Pruning a neuron in the MLP corresponds to removing the respective row and column from the weight and adjacency matrices of the graph. In the neural network model, this transfers to removing the $i$th row of a weight matrix $W^{(k)}$ as well as the $i$th column of $W^{(k+1)}$.
\par
The process of identifying which rows and columns to remove from $W^G$ (and $A^G$) using WPR requires operations on the large highly sparse $|V|\times |V|$ graph matrix, with exponentially more elements than the original MLP weight matrices. 
\par
However, as the only operation with $W^G$ (and $A^G$) in the WPR centrality measure is matrix-vector multiplication, the computation can be decomposed. This is based on the fact that all non-zero elements of $W^G$ are clearly defined, allowing for the following simplification

\begin{equation}
    W^G\bm x = \text{concat}((W^{(0)} \bm x^{(0)})^T, (W^{(1)} \bm x^{(1)})^T, \dots, (W^{(k)} \bm x^{(k)})^T)^T
    \label{equ:component_trick}
\end{equation}

where $\text{concat}()$ takes column vectors of different lengths as arguments and concatenates them along the first dimension in order to form a single large column vector.
In this component-wise setting, the directed weighted edges from layer $k-1$ to layer $k$ are associated with the weights of $W^{(k)}$. A simple clarifying example for the component-wise graph representation is given in \ref{apx:decomposition_example}. 

\subsection{Modified Weighted Page Rank}

Using the component-wise graph representation we compute importance scores based on a modified version of WPR as defined in \autoref{equ:mlp_wpr}. Motivated by the fact that we use uniform pruning we employ layer wise normalisation of the auxiliary information $\beta$ instead of global unit normalisation. To align with the component wise graph representation we use the individual weight matrix blocks and layer-wise importance score vectors. Note that $\gamma$ and $\theta$ are used as defined in \autoref{sec:wpr} and $\beta$ is used to include output activations as a norm over a representative calibration dataset, denoted as $\lVert X^{(k+1)}_j \rVert_2$. Here The latter is an important aspect as it is the only way that activation functions are taken into account by the scoring method. 

\begin{equation}
    \phi^{(k)}_i = \gamma \sum_j^{n^{(k)}}\Big( \theta \frac{|W^{(k)}_{ij}|}{\sum_l^{m^{(k)}} |W^{(k)}_{lj}|}+ (1-\theta)\frac{A^{(k)}_{ij}}{\sum_l^{m^{(k)}} A^{(k)}_{lj}} \Big) \phi^{(k-1)}_j + \frac{(1-\gamma) \lVert X^{(k+1)}_i \rVert_2}{\sum^{m^{(k)}}_l \lVert X^{(k+1)}_l \rVert_2}
    \label{equ:mlp_wpr}
\end{equation}

To assure convergence with the power iteration method \cite{herrmann1983iteration} to positive importance scores, we restrict elements of the recursive formulation to positive values which is done by using absolute weights.

\section{LLM-Rank Method}
\label{llmrank_method}

\llmrank extends the \mlprank pruning method to more complicated transformer-based architectures. The main difference is an updated graph representation which captures the structure of the transformer. Similar to \mlprank component-wise WPR is utilised for the computation of importance scores. As such \llmrank also is a post pruning strategy without weight updates which induces uniform structured sparsity. 
\par
The main challenge in the adaptation of MLP-Rank to transformers is the representation of the model architectures as a directed graph. This is particularly true as not all linear projections in the transformer are of the form input times weight matrix (plus bias) which is the implicit assumption of the graph representation so far. An example of this is the scaled dot product attention mechanism (\autoref{equ:attention}) which includes the multiplication of input projections with each other. 
\par
A straightforward way of dealing with this difficulty of representing attention is simply not to prune the MHA components in the transformer. Instead, we could view each FFN individually as an MLP and apply MLP-Rank pruning to it. This could already achieve significant speedups, as most of the parameters in a transformer reside in the FFNs components \citep{lin2022survey}.
\par 
A more sophisticated approach is to chain the individual FFN networks using the fact that the attention components only add incremental changes, due to the skip connections. This allows us to propose to structurally disregard the attention operation itself and only include its transformations to the information flow through the network based on calibration activations. This is similar to how activation functions are handled in the MLP networks. 
With this approach, we treat all FFNs as a single deep MLP and apply component-wise WPR centrality from \autoref{equ:mlp_wpr} for the computation of importance scores. The calibration activations are however still computed on the unchanged transformer model. This means the chained representation is only created for the computation of importance scores, which are then used to prune the original model.
\par
\begin{figure}[t]
\begin{center}
\includegraphics[width=\textwidth]{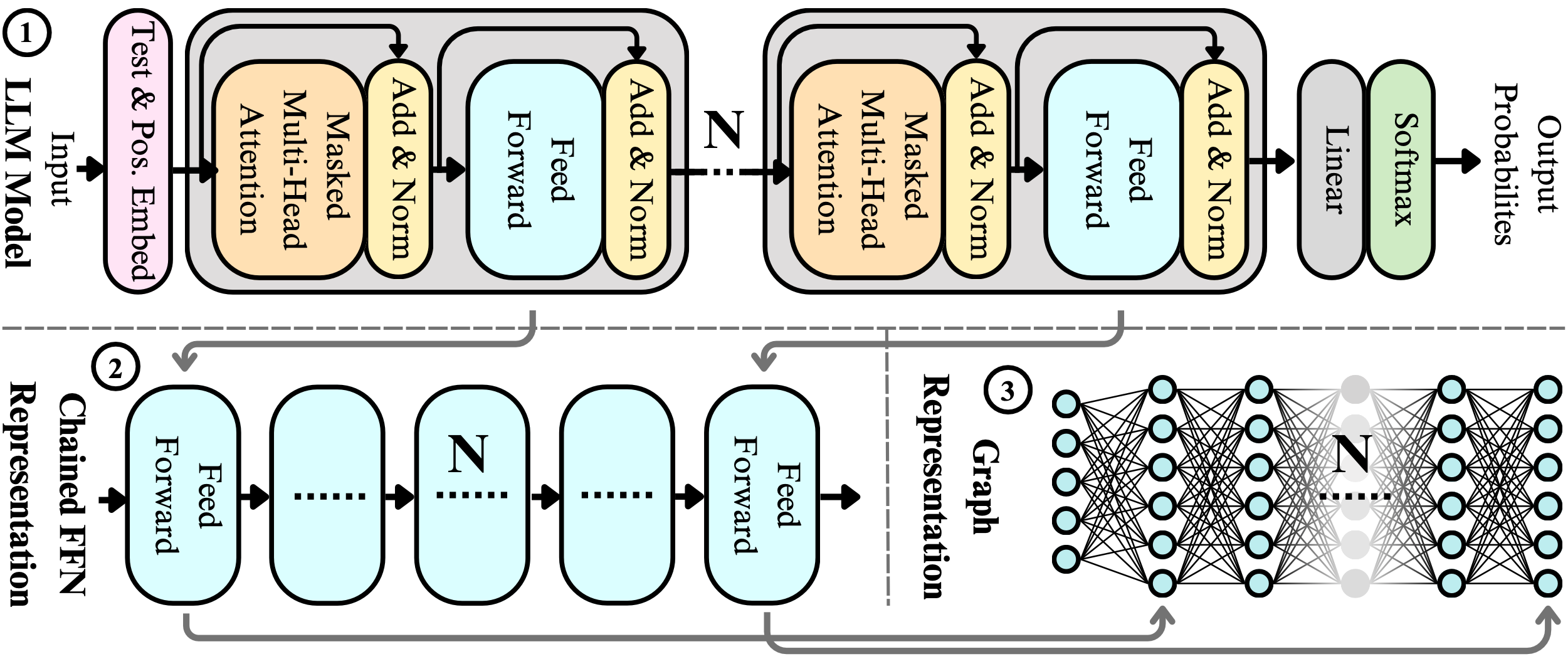}
\end{center}
\caption{Illustration of the proposed graph representation extraction from the LLM FFNs. Part (1) shows the full decoder-only transformer model from which the position-wise feed-forward networks are extracted and chained in part (2). The Graph representation of the resulting MLP network to which WPR is applied is shown in part (3). Note that the grey arrows represent the mapping from original architecture components, to the chained FFN, to the graph representation.}
\label{fig:llm_to_mlp}
\end{figure}

We call this method of representing decoder-only transformer models \llmrank. The extraction of the graph representation from the decoder-only LLM architecture is visualised in \autoref{fig:llm_to_mlp}.

A further improvement to the proposed method would be to include key-, value-, and query-weight matrices of the attention mechanism as appendices to the chain of FFNs. In particular, we can use the importance score vector of the previous FFN layer as the incoming importance to all three weight matrices. The implementation of this last aspect of pruning the attention weight matrices is further outlined in \autoref{apx:extentson_to_attention} but otherwise left for future research in this area.

\section{Experiments}
\label{experiments}

\subsection{MLP-Rank Accuracy Comparison}

The \mlprank method is evaluated on four different multilayer perceptrons with various sizes and trained on different datasets, namely MNIST \cite{lecun_mnist_2010}, Fashion-MNIST \cite{xiao_fashion_2017}, CIFAR10 \cite{krizhevsky_cifar_2012} and CIFAR100 \cite{krizhevsky_cifar_2012}. The models and the respective dataset they were trained on can be seen in \autoref{tab:mlprank_std}.

\begin{table}
    \centering
    \caption{Overview over the MLP models used for experimentation. Here number of in and out features is determined by number of features and number of classes of the dataset respectively.}
    \label{tab:mlp_models}
    \begin{tabular}{|l|l|l|l|l|l|}
        \hline
        Name & Dataset & \# Layers & \# In & \# Out & \# Parameters  \\
        \hline
        TwoLayerMLP & MNIST & 2 & 784 & 10 & 203,530  \\ 
        ThreeLayerMLP & Fashion-MNIST & 3 & 784 & 10 & 669,706 \\ 
        SixLayerMLP & CIFAR10 & 6 & 3072 & 10 & 2,069,130 \\ 
        TwelveLayerMLP & CIFAR100 & 12 & 3072 & 100 & 16,689,764 \\ 
        \hline
    \end{tabular}
\end{table}

Each model is trained until convergence using the ADAM optimiser \cite{kingma_adam_2017} with approximations of optimal hyperparameters for learning rate and weight decay, determined with a simple grid search. For the final evaluation we use 80\% of the data for training and 20\% for accuracy evaluation. 
\par
\autoref{tab:mlprank_avg_acc} shows the comparison of \mlprank pruning with random structured pruning, L1-Norm pruning and pruning based on layer-output activations. The table shows the average relative amount of top-1 and top-5 accuracy retention compared to the respective dense MLP across all four models. Standard deviations are shown in \autoref{tab:mlprank_std}.

\begin{table}
    \centering
    \footnotesize
    \caption{Average local pruning performance in terms of accuracy retention, for five amounts of local pruning, averaged across models. \textbf{Higher is better}.} 
    \label{tab:mlprank_avg_acc}
    \begin{tabular}{l|llllllllll}
        \midrule
        \hline
        Sparsity & \multicolumn{2}{c}{10\%} & \multicolumn{2}{c}{20\%} & \multicolumn{2}{c}{30\%} & \multicolumn{2}{c}{40\%} & \multicolumn{2}{c}{50\%} \\ 
        \hline
        Metric & Top-1 & Top-5 & Top-1 & Top-5 & Top-1 & Top-5 & Top-1 & Top5 & Top-1 & Top-5 \\ 
        \hline
        \rowcolor{lightgray}
        \mlprank & \textbf{91.69} & \textbf{96.36} & \textbf{78.70} & \textbf{91.55} & \textbf{66.29} & \textbf{85.87} & \textbf{54.52} & \textbf{78.35} & \textbf{44.99} & 71.93 \\
        Random & 85.39 & 94.04 & 72.39 & 88.07 & 53.91 & 79.77 & 44.86 & 72.57 & 32.90 & 61.45 \\
        L1-Norm & 90.27 & 95.62 & 76.95 & 89.83 & 63.95 & 83.31 & 51.98 & 75.65 & 41.28 & 69.50 \\
        Activation & 88.42 & 95.61 & 77.38 & 89.90 & 65.31 & 84.41 & 54.33 & 78.68 & 44.36 & \textbf{72.37} \\
        \midrule
        \hline
    \end{tabular}
\end{table}

We find that \mlprank outperforms all three baselines in terms of top-1 accuracy and for four out five sparsity ratios with top-5 accuracy. On average our method has a 13.64\% higher accuracy retention than random pruning, 3.42\% higher than L1-Norm pruning and 1.19\% higher than pruning based on layer-output activations. Overall it has a 6.09 \% higher accuracy retention than the three baselines per sparsity ratio. More granular results can be found in \ref{apx:mlp_experiments}.

\subsection{LLM-Rank Zero-Shot Comparison}
\label{sec:zero_shor_performance}

We evaluate \llmrank pruning on the LLaMA architecture, in particular, the Open-LLaMa-3b-v2 model provided by \cite{openlm2023openllama}. The performance of the pruned model is evaluated on six popular LLM benchmarks using the EleutherAI LM Harness \cite{eval-harness}, namely ARC-Challenge, ARC-easy, HellaSwag, OpenBookQA, PIQA and WinoGrande.
\par
\llmrank pruning is compared with two baselines, the naive L1-Norm pruning approach and the state-of-the-art method SliceGPT \citep{slice2024} on 5 different levels of sparsity. The calibration activations are computed on random samples from the C4 dataset \citep{raffel2023exploringlimitstransferlearning}.
The experiments were run on an Amazon EC2 p3.16xlarge instance with 8 V100 GPUs, where each pruning run with \llmrank had an average wall time of only 53.84 seconds with a standard deviation of 1.20 seconds. 
\par
\autoref{tab:zero_shot_acc} shows the zero-shot accuracy comparison in percent between \llmrank using chained FFNs and two popular baselines. The first baseline is L1-Norm pruning which is similar to magnitude pruning and often times used as a naive but strong baseline for pruning. The second one is SliceGPT which was introduced by \cite{slice2024} and is one of the current state-of-the-art methods for structured pruning of LLMs. They are a good comparison point to \llmrank as neither use post pruning tuning which leads to the noticeable gap between dense and pruned model performance for higher sparsity ratio. 

\begin{table}[ht]
    \centering
    \caption{Zero-Shot accuracy (\%) comparison on Open-LLaMa-3b-v2 of \llmrank and two baseline methods for 6 popular LLM benchmarks. Where\llmrank outperforms both baselines for four out of five sparsity ratios on average across the benchmarks. \textbf{Higher is better}.} 
    \label{tab:zero_shot_acc}
    \begin{tabular}{l|l|rrrrrrr}
        \midrule
        \hline
         &  & ARC-c & ARC-e & Hella- & Open- & PIQA & Wino- & Mean \\
        Sparsity & Score &  &  & Swag & BookQA &  & Grande & \\
        \hline
        00\% & Dense & 33.53 & 67.72 & 52.19 & 26.20 & 76.93 & 63.14 & 53.29 \\
            \hline
            \rowcolor{lightgray}
            \multirow[t]{3}{*}{05\%} & LLM-Rank & \bf{28.50} & \bf{63.38} & \bf{46.84} & \bf{22.80} & \bf{73.88} & \bf{61.72} & \bf{49.52} \\
                                     & SliceGPT &     23.46  &     50.93  &     41.23  &     19.40  &     70.29  &     55.80  &     43.52  \\
                                     & L1-Norm  &     22.61  &     25.88  &     25.90  &     17.20  &     54.46  &     49.88  &     32.66  \\
            \hline
            \rowcolor{lightgray}
            \multirow[t]{3}{*}{10\%} & LLM-Rank & \bf{25.26} & \bf{52.31} & \bf{40.42} &     18.20  & \bf{68.77} & \bf{56.99} & \bf{43.66} \\
                                     & SliceGPT &     20.65  &     32.87  &     30.38  & \bf{18.40} &     59.85  &     53.67  &     35.97  \\
                                     & L1-Norm  &     21.67  &     25.80  &     25.91  &     18.20  &     53.05  &     49.09  &     32.29  \\
            \hline
            \rowcolor{lightgray}
            \multirow[t]{3}{*}{15\%} & LLM-Rank & \bf{21.42} & \bf{39.73} & \bf{33.45} &     14.00  & \bf{62.08} &     50.28  & \bf{36.83} \\
                                     & SliceGPT &     20.05  &     27.31  &     27.07  &     15.40  &     54.68  & \bf{51.70} &     32.70  \\
                                     & L1-Norm  &     19.88  &     26.14  &     25.64  & \bf{16.20} &     51.80  &     49.64  &     31.55  \\
            \hline
            \rowcolor{lightgray}
            \multirow[t]{3}{*}{20\%} & LLM-Rank &     20.05  & \bf{30.98} & \bf{27.79} &     16.20  & \bf{55.82} & \bf{51.38} & \bf{33.70} \\
                                     & SliceGPT & \bf{20.73} &     27.69  &     26.96  &     12.60  &     54.46  &     50.28  &     32.12  \\
                                     & L1-Norm  &     19.20  &     26.22  &     25.97  & \bf{16.40} &     53.43  &     48.78  &     31.67  \\
            \hline
            \rowcolor{lightgray}
            \multirow[t]{3}{*}{30\%} & LLM-Rank &     20.05  &     26.26  &     25.87  &     13.40  &     51.20  &     50.20  &     31.16  \\
                                     & SliceGPT & \bf{21.76} & \bf{27.61} & \bf{26.51} &     12.80  & \bf{53.48} & \bf{51.46} & \bf{32.27} \\
                                     & L1-Norm  &     21.33  &     24.58  &     26.05  & \bf{16.00} &     53.21  &     50.36  &     31.92  \\
            \hline
            \rowcolor{lightgray}
            \multirow[t]{3}{*}{40\%} & LLM-Rank & \bf{22.70} &     25.29  &     25.71  & \bf{17.00} &     53.26  &     50.20  & \bf{32.36} \\
                                     & SliceGPT &     20.82  & \bf{27.48} & \bf{25.95} &     12.00  & \bf{53.43} &     50.12  &     31.63  \\
                                     & L1-Norm  &     20.56  &     24.83  &     25.66  &     16.00  &     52.61  & \bf{50.91} &     31.76  \\
        \midrule
        \hline
    \end{tabular}
\end{table}
\begin{wrapfigure}{r}{0.45\textwidth}
    \vspace{-10pt}
    \begin{center}
        \includegraphics[width=0.44\textwidth]{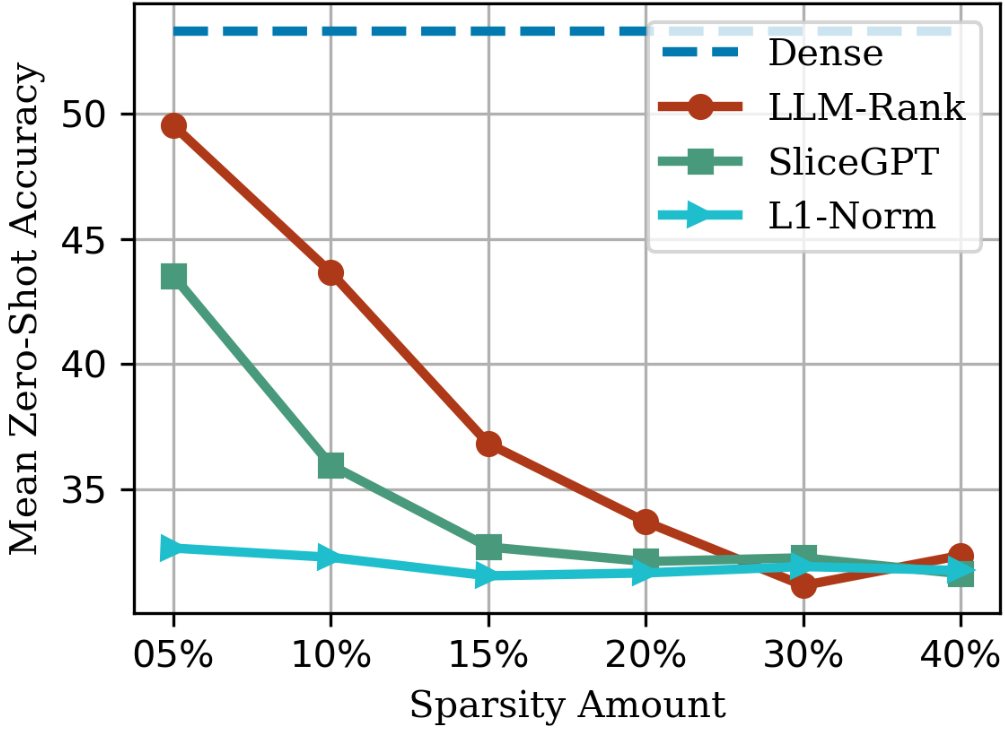}
    \end{center}
    \vspace{-10pt}
    \caption{Mean zero-shot accuracy for each method across all five sparsity ratios.}
    \label{fig:results_plot}
    \vspace{-10pt}
\end{wrapfigure}

The comparison shows that \llmrank outperforms the two baselines for lower sparsity ratios and has comparable performance for higher sparsity ratios. In particular, it has the highest average accuracy for four out of five sparsity ratios as visualised in \autoref{fig:results_plot}. In particular, \llmrank on average has a 18.25 \% higher accuracy per sparsity ratio than L1-Norm pruning and an 8.60 \% higher accuracy accuracy per sparsity ratio than SliceGPT. A table with the standard deviations for \autoref{tab:zero_shot_acc} can be found in \ref{sec:zero_shot_std}.

Note that the difference in average performance between \llmrank and the two baselines seems to be negatively correlated with the sparsity ratios. We hypothesise that this may be partly attributed to the fact that there are only a few fully redundant structures which our method identifies very effectively. However, when moving to higher sparsity ratios the relative share of non-redundant parameters pruned decreases. This may lead to the diminished difference in performance between the pruning methods. 
\par
The performance decrease of our method for higher sparsity ratios may also be due to the fact that we are only pruning the FFN weight matrices. This is not the case for SliceGPT which also prunes parameters in the MHA component of the LLM allowing for a more streamlined information flow. For example, \llmrank has to prune around 80\% of parameters in the FFN components to reach an overall sparsity of 40\% in the LLM leading to tighter information bottlenecks compared to SliceGPT which simply prunes 40\% of every individual weight matrix.
\par
From this we can firstly infer that our method is powerful at selecting redundant elements, as even with the mentioned information bottlenecks it outperforms the two baselines, and secondly that an extension to the attention weight matrices is expected to lead to even greater performance.

\subsection{Inference Speedup}

The aim of pruning as a way of inference optimisation in LLMs, is to speed up computations and reduce memory consumption by removing parameter dropping and operations associated with them. Structured pruning as performed by \llmrank leads to real-world speedups without the need for specially designed hardware or software. This is in opposition to unstructured or semi-structured sparsity which requires specialised hardware to achieve real speedups. 
\par
When pruning the same share of nodes per layer, the relationship between the amount of pruned nodes, number of parameters, number of FLOPs and theoretical speedup is straightforward. Here pruning 50\% of neurons per layer corresponds to halving the number of parameters in the network as well as the number of FLOPs which then leads to a two times speedup in inference latency. This direct translation of sparsity ratio to inference speedup is also expected for our method, but not empirically investigated as part of this paper.

\section{Conclusion}
\label{conclusion}

Motivated by the ever-growing sizes of LLMs and the need for effective pruning methods, we propose \mlprank and \llmrank, a graph theoretical approach to pruning both MLPs and LLMs. In particular we propose a way of creating a weighted directed acyclical graph representation from multilayer perceptrons as well as decoder-only transformers. We then apply a modified version of weighted PageRank to this graph and use it for the computation of importance scores based on which we perform structured uniform pruning. We show that \mlprank outperforms random, L1-Norm and activation based pruning with an on average 6.09 \% higher accuracy per sparsity ratio. The \llmrank method is compared to L1-Norm pruning and the state-of-the-art method SliceGPT, outperforming both for lower sparsity ratios and approaching their performance for highter sparsity ratios for six popular LLM benchmarks on the Open-LLaMa-3b-v2 model. On average it has a 13.42 \% higher accuracy compared per sparsity ratio compared two the two baselines. 
\par
Two important next steps, left to future research are the replication of our results on different LLM architectures and model size as well as further work related to the graph representation of the attention mechanism. Solving this would allow for more extensive pruning of all weight matrices that are part of a transformer model and possibly further improve pruning performance.

\bibliography{iclr2025_conference}
\bibliographystyle{iclr2025_conference}

\appendix
\section{MLP-Rank Pruning Representation}

\subsection{Matrix Decomposition Example}
\label{apx:decomposition_example}

This is a simple example of the graph representation of a two-layer MLP, without nonlinearities and activations and weight matrices

\begin{equation}
    W^{(0)}=\begin{pmatrix} 1 & 2 \\ 3 & 4 \end{pmatrix} \text{ and } W^{(1)}=\begin{pmatrix} 5 & 6 \\ 7 & 8 \end{pmatrix}.
    \label{equ:methodology_graph_representation_example_mlp}
\end{equation}

Here both $W^{(1)}$ and $W^{(2)}$ have the dimension $2 \times 2$. This means that the MLP takes an input vector $x^{(1)}$ of length $2$ and outputs a vector of length $2$. The graph associated with this network has $6$ nodes. Two surrogate nodes are associated with the input vector, two with the hidden layer and two corresponding to the final output layer. The weight, as well as adjacency matrices, are defined in \autoref{equ:methodology_graph_representation_example_graph}.

\begin{equation}
    W^{G}=\begin{pmatrix} 
        0 & 0 & 0 & 0 & 0 & 0 \\ 
        0 & 0 & 0 & 0 & 0 & 0 \\ 
        1 & 2 & 0 & 0 & 0 & 0 \\ 
        3 & 4 & 0 & 0 & 0 & 0 \\
        0 & 0 & 5 & 6 & 0 & 0 \\
        0 & 0 & 7 & 8 & 0 & 0 \\
    \end{pmatrix} 
    \text{ and } 
    A^{G}=\begin{pmatrix} 
        0 & 0 & 0 & 0 & 0 & 0 \\ 
        0 & 0 & 0 & 0 & 0 & 0 \\ 
        1 & 1 & 0 & 0 & 0 & 0 \\ 
        1 & 1 & 0 & 0 & 0 & 0 \\
        0 & 0 & 1 & 1 & 0 & 0 \\
        0 & 0 & 1 & 1 & 0 & 0 \\
    \end{pmatrix}
    \text{.}
    \label{equ:methodology_graph_representation_example_graph}
\end{equation}

\begin{figure}[ht]
    \centering
    \includegraphics[width=0.6\textwidth]{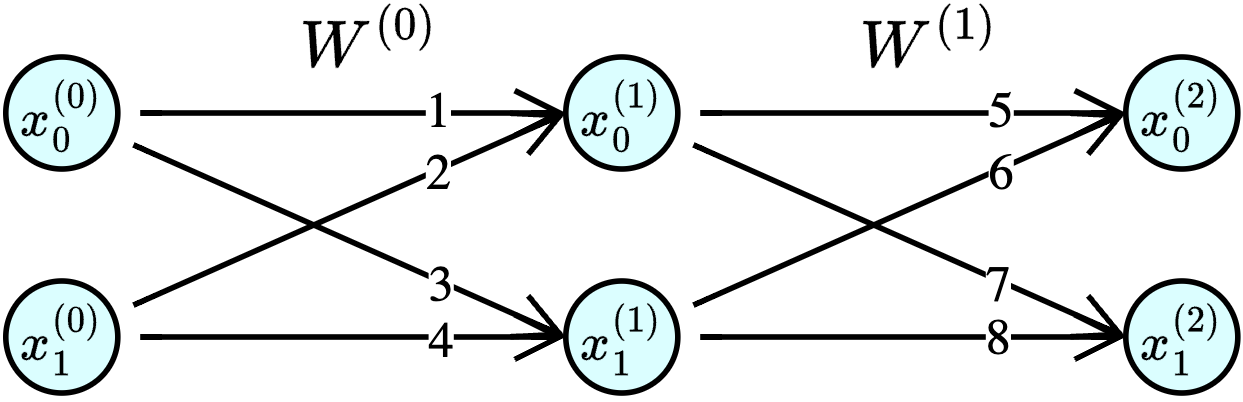}
    \caption{Example of a two-layer MLP graph representation \citep{hoffmann2024mlprank}.}
    \label{fig:methodology_graph_representation_graph_example_mlp}
\end{figure}

A visualisation of the weight matrix of the graph representation can be found in \autoref{fig:methodology_graph_representation_graph_example_mlp}. It clearly shows the structure of weights in the weight matrix providing an intuition for how the decomposition outlined in \autoref{equ:graph_representation} is possible.

\subsection{Graph Representation Weight Matrix}
\label{apx:graph_representation}

\begin{figure}[ht]
    \begin{center}
    \includegraphics[width=0.6\textwidth]{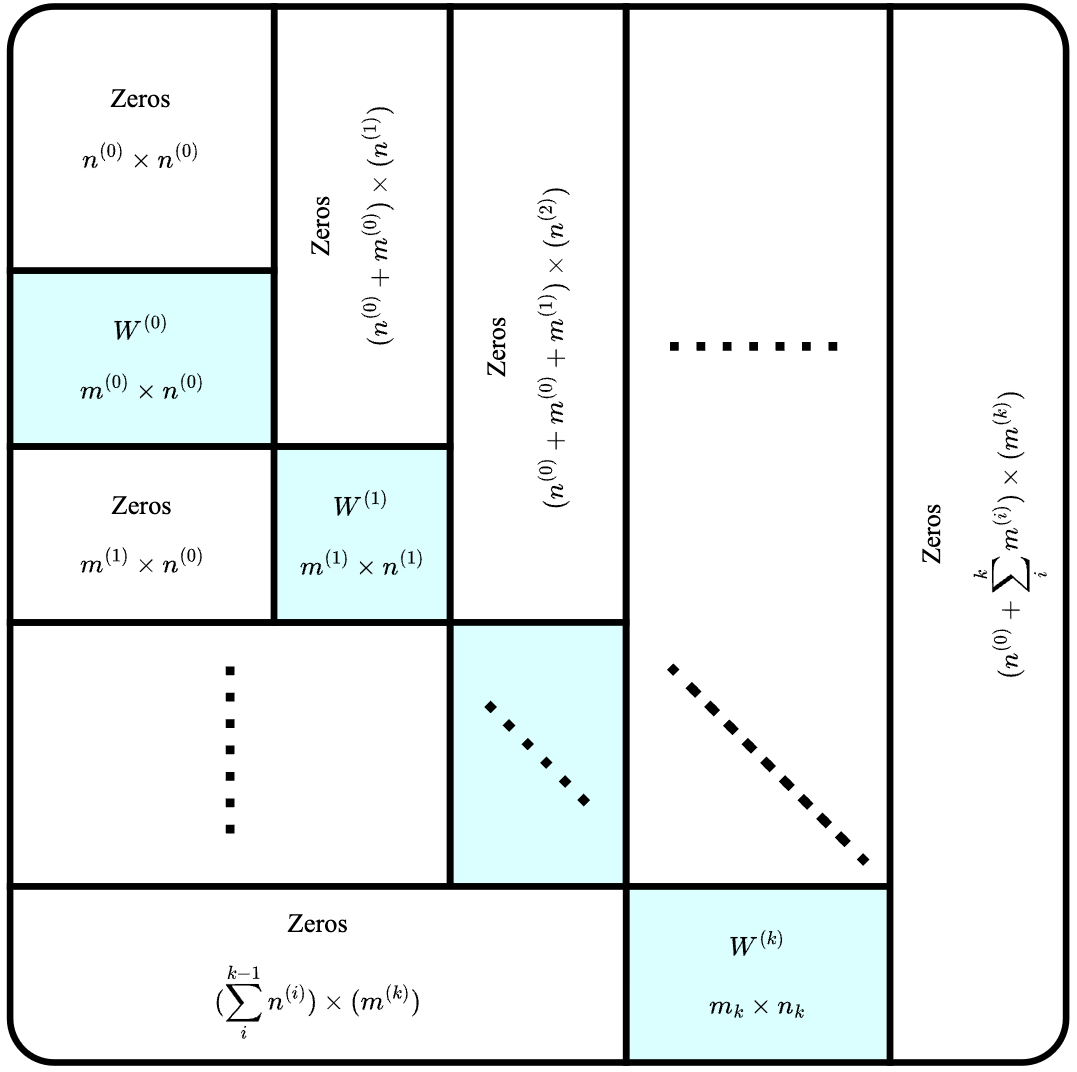}
    \end{center}
    \caption{Weight matrix of the graph representation \citep{hoffmann2024mlprank}.}
    \label{fig:graph_representation}
\end{figure}

\autoref{fig:graph_representation} provides a good overview of the structure of $W^G$, showing how output dimensions $m^{(k)}$ of one layer have to match the input dimensions $n^{(k+1)}$ of the following layer. 

\section{Experiment Results}

\subsection{MLP-Rank Experiments}
\label{apx:mlp_experiments}

\autoref{tab:mlprank_std} documents the standard deviations corresponding to the aggregated comparison of retained accuracy for different scoring functions as presented in \autoref{tab:mlprank_avg_acc}.
The relatively high standard deviation values of accuracy retention are caused by the large difference between models across which results are aggregated in this table. They are not an indication of high variance in the pruning methods itself which are deterministic.

\begin{table}
    \centering
    \footnotesize
    \caption{Local Pruning Performance. The standard deviation for the performance retention in percent for different amounts of sparsity.} 
    \label{tab:mlprank_std}
    \begin{tabular}{l|llllllllll}
        \midrule
        \hline
        Sparsity & \multicolumn{2}{c}{10\%} & \multicolumn{2}{c}{20\%} & \multicolumn{2}{c}{30\%} & \multicolumn{2}{c}{40\%} & \multicolumn{2}{c}{50\%} \\ 
        \hline
        Scoring & Top-1 & Top-5 & Top-1 & Top-5 & Top-1 & Top-5 & Top-1 & Top5 & Top-1 & Top-5 \\ 
        \hline
        \rowcolor{lightgray}
        \mlprank & 5.17 & 5.17 & 10.67 & 11.74 & 16.26 & 18.54 & 20.89 & 25.02 & 19.15 & 28.4 \\
        Random & 8.99 & 8.37 & 14.38 & 16.18 & 16.01 & 22.91 & 18.11 & 28.17 & 16.74 & 27.27 \\
        L1-Norm & 6.37 & 5.68 & 11.86 & 13.15 & 19.04 & 22.13 & 22.04 & 29.09 & 20.33 & 31.72 \\
        Activation & 6.98 & 6.42 & 13.45 & 12.99 & 18.82 & 19.66 & 20.63 & 25.07 & 19.37 & 28.14 \\
        \midrule
        \hline
    \end{tabular}
\end{table}

In the following, the comparison of different scoring measures will be further broken down by comparing the raw accuracy results for each of the four MLP models individually.

\begin{table}
    \centering
    \footnotesize
    \caption{TwoLayerMLP pruning performance in terms of accuracy retention, compared for five amounts of local pruning, with the winning scoring function of each column marked in bold.} 
    \begin{tabular}{l|llllllllll}
        \midrule
        \hline
        Sparsity & \multicolumn{2}{c}{10\%} & \multicolumn{2}{c}{20\%} & \multicolumn{2}{c}{30\%} & \multicolumn{2}{c}{40\%} & \multicolumn{2}{c}{50\%} \\ \hline
        Scoring & Top-1 & Top-5 & Top-1 & Top-5 & Top-1 & Top-5 & Top-1 & Top5 & Top-1 & Top-5 \\ 
        \hline
        \rowcolor{lightgray}
        \mlprank & \textbf{95.16} & \textbf{100.01} & \textbf{87.30} & \textbf{99.98} & \textbf{78.57} & 99.83 & \textbf{69.31} & 96.63 & \textbf{58.22} & 90.57 \\
        Random & 93.22 & 99.98 & 83.81 & 99.92 & 69.41 & 99.6 & 62.02 & 97.14 & 53.28 & 89.96 \\
        L1-Norm & 92.41 & 99.98 & 83.65 & 99.92 & 74.20 & 99.82 & 68.49 & \textbf{99.08} & 52.39 & \textbf{90.74} \\
        Activation & 95.04 & \textbf{100.01} & 85.23 & 99.97 & 73.60 & \textbf{99.84} & 66.05 & 97.56 & 55.91 & 90.22 \\
        \midrule
        \hline
    \end{tabular}
\end{table}

\begin{table}
    \centering
    \footnotesize
    \caption{ThreeLayerMLP pruning performance in terms of accuracy retention, compared for five amounts of local pruning. with the winning scoring function of each column marked in bold.} 
    \begin{tabular}{l|llllllllll}
        \midrule
        \hline
        Sparsity & \multicolumn{2}{c}{10\%} & \multicolumn{2}{c}{20\%} & \multicolumn{2}{c}{30\%} & \multicolumn{2}{c}{40\%} & \multicolumn{2}{c}{50\%} \\ \hline
        Scoring & Top-1 & Top-5 & Top-1 & Top-5 & Top-1 & Top-5 & Top-1 & Top5 & Top-1 & Top-5 \\ 
        \hline
        \rowcolor{lightgray}
        \mlprank & 94.07 & 99.96 & 83.58 & 99.62 & 79.32 & 98.74 & 73.05 & 98.83 & 61.49 & 97.69 \\
        Random & 92.33 & 99.87 & 82.11 & 98.85 & 64.27 & 96.02 & 54.79 & 92.41 & 30.79 & 71.03 \\
        L1-Norm & \textbf{98.27} & 99.75 & 88.75 & 99.28 & 80.64 & 98.68 & 69.65 & 97.52 & 58.76 & 96.48 \\
        Activation & 94.08 & \textbf{99.97} & \textbf{89.86} & \textbf{99.97} & \textbf{84.92} & \textbf{99.84} & \textbf{75.16} & \textbf{99.55} & \textbf{64.48} & \textbf{99.23} \\
        \midrule
        \hline
    \end{tabular}
\end{table}

\begin{table}
    \centering
    \footnotesize
    \caption{SixLayerMLP pruning performance, in terms of accuracy retention, compared for five amounts of local pruning, with the winning scoring function of each column marked in bold.}
    \begin{tabular}{l|llllllllll}
        \midrule
        \hline
        Sparsity & \multicolumn{2}{c}{10\%} & \multicolumn{2}{c}{20\%} & \multicolumn{2}{c}{30\%} & \multicolumn{2}{c}{40\%} & \multicolumn{2}{c}{50\%} \\ 
        \hline
        Scoring & Top-1 & Top-5 & Top-1 & Top-5 & Top-1 & Top-5 & Top-1 & Top5 & Top-1 & Top-5 \\ 
        \hline
        \rowcolor{lightgray}
        \mlprank & \textbf{94.77} & \textbf{97.93} & \textbf{83.51} & \textbf{95.12} & 68.06 & \textbf{90.54} & \textbf{55.56} & \textbf{81.33} & \textbf{47.13} & \textbf{74.52} \\
        Random & 85.24 & 96.57 & 75.58 & 93.10 & 54.09 & 81.60 & 47.96 & 74.76 & 40.10 & 68.35 \\
        L1-Norm & 89.81 & 96.71 & 77.87 & 92.50 & \textbf{69.23} & 89.10 & 54.57 & 78.80 & 47.16 & 74.39 \\
        Activation & 86.96 & 97.87 & 79.48 & 91.44 & 68.29 & 86.23 & 55.40 & 80.39 & 43.69 & 73.77 \\
        \midrule
        \hline
    \end{tabular}
\end{table}

\begin{table}
    \centering
    \footnotesize
    \caption{TwelveLayerMLP pruning performance, in terms of accuracy retention, compared for five amounts of local pruning, with the winning scoring function of each column is marked in bold.}
    \begin{tabular}{l|llllllllll}
        \midrule
        \hline
        Sparsity & \multicolumn{2}{c}{10\%} & \multicolumn{2}{c}{20\%} & \multicolumn{2}{c}{30\%} & \multicolumn{2}{c}{40\%} & \multicolumn{2}{c}{50\%} \\ \hline
        Scoring & Top-1 & Top-5 & Top-1 & Top-5 & Top-1 & Top-5 & Top-1 & Top5 & Top-1 & Top-5 \\ 
        \hline
        \rowcolor{lightgray}
        \mlprank & \textbf{82.77} & \textbf{87.52} & \textbf{60.42} & \textbf{71.49} & \textbf{39.21} & \textbf{54.36} & 20.14 & 36.61 & 13.14 & 24.93 \\
        Random & 70.77 & 79.75 & 48.06 & 60.4 & 27.88 & 41.84 & 14.7 & 25.97 & 7.41 & 16.47 \\
        L1-Norm & 80.6 & 86.05 & 57.51 & 67.61 & 31.72 & 45.66 & 15.23 & 27.21 & 6.79 & 16.37 \\
        Activation & 77.61 & 84.59 & 54.97 & 68.23 & 34.42 & 51.74 & \textbf{20.71} & \textbf{37.23} & \textbf{13.34} & \textbf{26.27} \\
        \midrule
        \hline
    \end{tabular}
\end{table}

\subsection{Zero-Shot Accuracy Standard Deviation}
\label{sec:zero_shot_std}
    
The table below shows the standard deviations produced by the LM-Evaluation Harness for the zero shot accuracy comparison between our method, L1-Norm pruning and SliceGPT. It is noteworthy that for the latter method, standard deviations in the benchmarking results are an order of magnitude larger than for L1-Norm pruning and \llmrank. 

\begin{table}[ht]
    \centering
    \caption{Standard deviations for the zero-shot accuracy (\%) comparison between \llmrank and the two baselines, namely L1-Norm pruning and SliceGPT.} 
    \label{tab:zero_shot_std}
    \begin{tabular}{llrrrrrrr}
        \midrule
        \hline
         &  & ARC-c & ARC-e & Hella- & Open- & PIQA & Wino- & Mean \\
        Sparsity & Score &  &  & Swag & BookQA &  & Grande & \\
        \hline
        00\% & Dense & 1.38 & 0.96 & 0.50 & 1.97 & 0.98 & 1.36 & 1.19 \\
        \hline
        \rowcolor{lightgray}
        \multirow[t]{3}{*}{05\%} & L1-Norm & 1.22 & 0.90 & 0.44 & 1.69 & 1.16 & 1.41 & 1.14 \\
        & LLM-Rank & 1.27 & 1.02 & 0.49 & 1.73 & 1.08 & 1.39 & 1.16 \\
        & SliceGPT & 1.24 & 1.03 & 0.49 & 1.77 & 1.07 & 1.40 & 1.16 \\
        \hline
        \rowcolor{lightgray}
        \multirow[t]{3}{*}{10\%} & L1-Norm & 1.20 & 0.90 & 0.44 & 1.73 & 1.16 & 1.41 & 1.14 \\
        & LLM-Rank & 1.32 & 0.99 & 0.50 & 1.88 & 1.02 & 1.37 & 1.18 \\
        & SliceGPT & 1.18 & 0.96 & 0.46 & 1.73 & 1.14 & 1.40 & 1.15 \\
        \hline
        \rowcolor{lightgray}
        \multirow[t]{3}{*}{15\%} & L1-Norm & 1.17 & 0.90 & 0.44 & 1.65 & 1.17 & 1.41 & 1.12 \\
        & LLM-Rank & 1.17 & 0.90 & 0.44 & 1.52 & 1.17 & 1.41 & 1.10 \\
        & SliceGPT & 1.17 & 0.91 & 0.44 & 1.62 & 1.16 & 1.40 & 1.12 \\
        \hline
        \rowcolor{lightgray}
        \multirow[t]{3}{*}{20\%} & L1-Norm & 1.15 & 0.90 & 0.44 & 1.66 & 1.16 & 1.40 & 1.12 \\
        & LLM-Rank & 1.17 & 0.95 & 0.45 & 1.65 & 1.16 & 1.40 & 1.13 \\
        & SliceGPT & 1.18 & 0.92 & 0.44 & 1.49 & 1.16 & 1.41 & 1.10 \\
        \hline
        \rowcolor{lightgray}
        \multirow[t]{3}{*}{30\%} & L1-Norm & 1.20 & 0.88 & 0.44 & 1.64 & 1.16 & 1.41 & 1.12 \\
        & LLM-Rank & 1.20 & 1.00 & 0.47 & 1.55 & 1.13 & 1.41 & 1.13 \\
        & SliceGPT & 1.21 & 0.92 & 0.44 & 1.50 & 1.16 & 1.40 & 1.10 \\
        \hline
        \rowcolor{lightgray}
        \multirow[t]{3}{*}{40\%} & L1-Norm & 1.18 & 0.89 & 0.44 & 1.64 & 1.16 & 1.41 & 1.12 \\
        & LLM-Rank & 1.22 & 0.89 & 0.44 & 1.68 & 1.16 & 1.41 & 1.13 \\
        & SliceGPT & 1.19 & 0.92 & 0.44 & 1.45 & 1.16 & 1.41 & 1.09 \\
        \midrule
        \hline
    \end{tabular}
\end{table}

\section{Extension to Attention}
\label{apx:extentson_to_attention}

\begin{figure}[ht]
    \centering
    \includegraphics[width=\linewidth]{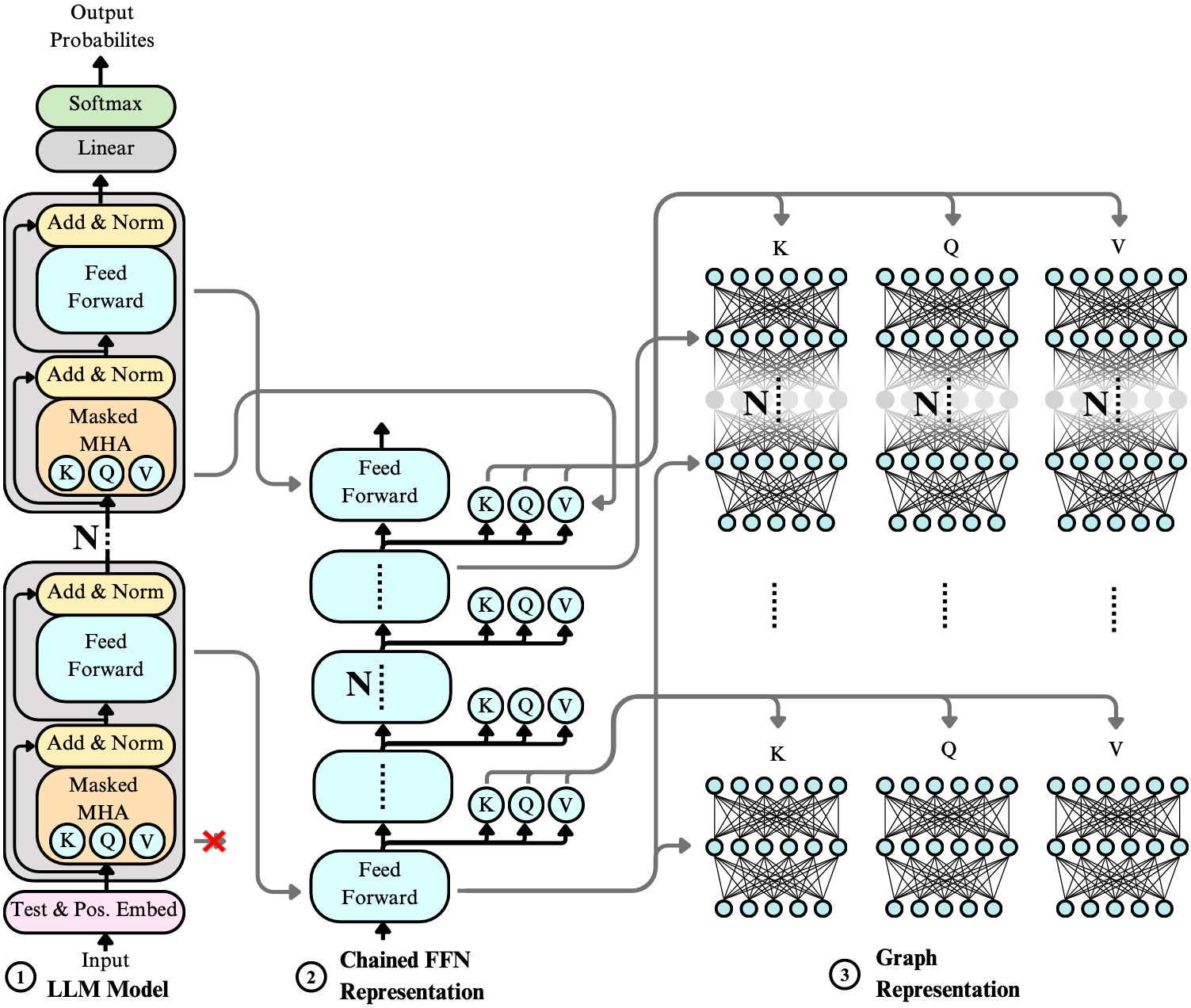}
    \caption{Exemplary illustration for the creation of an LLM graph representation including key-, value-, and query matrices. Note that the grey arrows represent the mapping from original architecture components, to the chained FFN, to the graph representation.}
    \label{fig:extentson_to_attention}
\end{figure}

The proposed \llmrank can be extended to include key-, value-, and query matrices from the multi-head attention (MHA) components for pruning as outlined in \autoref{llmrank_method}. \autoref{fig:extentson_to_attention} shows how this could be achieved in two steps: (1) the extraction and chaining of the FFNs with key-, value-, and query matrices as appendices to this chain. (2) the creation of multiple MLP networks each consisting of the chain of previous FFNs and ending in one of the three attention matrices. 

As each layer is pruned based on information from its layer as well as previous layers, additionally to the full chained FFN, a series of smaller MLPs are created for each of the attention weight matrices in the LLM. These smaller networks include all FFNs prior to the attention weight matrix which itself always represents the final layer. The component-wise WPR is then applied to each individual network for the importance score computation. To reduce computations, the importance scores of the full chained FFNs network can be cashed and reused for the computation of the key-, value-, and query matrices.
The implementation and evaluation of this extension to \llmrank is left to future research. 
\end{document}